\documentclass{ceurart}
\usepackage{graphicx}
\usepackage{booktabs}
\usepackage{amsmath}
\usepackage{url}
\usepackage{microtype}
\usepackage{subcaption}
\usepackage{placeins}
\title{Toward Sustainable Agentic AI: Carbon Tradeoffs Across Iterative Self-Refinement
  and Tool-Augmented Coding}

  \author[]{Boris Ruf}[suffix=\textsuperscript{\ensuremath{\dagger}},email=boris.ruf@axa.com]
  \author[]{Adrien {de Mailly Nesle}}[suffix=\textsuperscript{\ensuremath{\dagger}},email=adrien.demaillynesle@axa.com]
  \author[]{Marcin Detyniecki}[email=marcin.detyniecki@axa.com]
  \address{AXA AI Research}
  \nonumnote{\textsuperscript{\ensuremath{\dagger}} Equal contribution.}

\begin{document}
\hypersetup{urlcolor=blue,breaklinks=true,citecolor=black}
\begin{abstract}
Agentic language model systems can improve performance either by iterating on their own outputs or by using external tools in grounded environments. While these strategies can reduce error rates and substitute for larger models in some settings, their environmental cost remains relatively underexplored. We present two complementary empirical studies that estimate the energy and carbon footprint of agentic AI across settings of increasing complexity. Study~1 evaluates iterative verbal self-refinement: GPT-4.1-mini at $k = 1$--$5$ steps versus a GPT-4.1 single-pass baseline on two MATH benchmark splits. Study~2 evaluates tool-augmented agentic coding: GPT-4.1-mini, Claude Opus~4.6, and GPT-5.3-codex on SWE-bench Lite, where agents operate inside Docker-isolated environments with access to code execution and file editing tools. In Study~1, $k=3$ refinements close a 16-percentage-point accuracy gap to within 1~point, but nearly triple the carbon footprint relative to the cheap single-pass baseline while remaining below the GPT-4.1 single-pass baseline under the current token-based carbon scenario. In Study~2, GPT-5.3-codex achieves the highest resolution rate in this sample (81.2\%), ahead of Claude Opus~4.6 (68.9\%) and GPT-4.1-mini (25.0\%), but resolved-task-normalized efficiency remains mixed: GPT-4.1-mini is lowest on estimated carbon and cost per resolved task, while GPT-5.3-codex delivers substantially higher task success. All carbon estimates cover output-token inference only; prompt tokens, which exceed 99\% of tokens in Study~2, and tool execution are not charged, so the figures are lower bounds on the full agentic footprint.
\end{abstract}


\conference{GREEN-AI Workshop at CIKM 2026}

\maketitle

\section{Introduction}

Large language model (LLM) agents increasingly rely on two broad strategies
to improve performance: iterative self-refinement and tool-augmented
interaction. In self-refinement, the model generates an initial response,
critiques it, and revises it over multiple rounds
\cite{madaan2023selfrefine,shinn2023reflexion}. In tool-augmented settings,
agents use code execution, file editing, and test runners to ground decisions
in external feedback \cite{yao2023react,openhands2024}.
Both strategies can improve task outcomes and, in some settings, allow smaller
models to approach the quality of larger ones \cite{yao2023react,chavan2024faster}.

From a \emph{sustainability} perspective, both framings are incomplete.
Self-refinement does not eliminate cost---it shifts it from model parameters
to API calls and execution time.
Each additional refinement step or tool invocation consumes prompt tokens,
completion tokens, compute time, and energy.
Understanding how quickly the carbon footprint compounds as a function of
refinement depth, agent architecture, and model choice is essential for
responsible deployment of agentic systems.

This paper makes contributions in the following fields:
\begin{itemize}
  \item \textbf{Iterative Self-Refinement} (Study~1). A controlled,
        forced-$k$ evaluation of GPT-4.1-mini ($k = 1$--$5$) against a
        GPT-4.1 single-pass baseline on MATH Level~3--4 and Level~4--5,
        with per-call token, energy, and carbon tracking.
  \item \textbf{Tool-Augmented Agentic Coding} (Study~2). A carbon
        analysis of a tool-augmented agentic coding benchmark:
        GPT-4.1-mini, Claude Opus~4.6, and GPT-5.3-codex on
        SWE-bench Lite \cite{jimenez2024swebench} under Docker isolation,
      estimating the LLM-side inference footprint of real-world software
      engineering tasks.
\end{itemize}

\section{Related Work}

\paragraph{Self-refinement in LLM agents.}
Self-Refine was introduced in \cite{madaan2023selfrefine}, demonstrating that iterative self-feedback improves performance
across a range of tasks.
Reflexion \cite{shinn2023reflexion} frames refinement as verbal reinforcement
learning.
Both lines of work focus on accuracy gains, including on mathematical reasoning tasks \cite{jiang2024llms}; energy costs are not reported.
ReAct \cite{yao2023react} extends agents to use tools, interleaving reasoning
and acting in a grounded loop.

\paragraph{Agentic software engineering.}
SWE-bench \cite{jimenez2024swebench} introduced a realistic benchmark where
agents must resolve real GitHub issues by editing source code and passing
repository tests.
OpenHands \cite{openhands2024} provides an open agent framework for this
setting, running each task in a Docker-isolated workspace with access to bash,
file editing, and browsing tools.
AutoAgents \cite{chen2024autoagents} demonstrates how agents can be generated
dynamically to coordinate specialised tools across complex multi-step tasks.
No prior work has characterised the carbon cost of these agentic coding
pipelines.

\paragraph{Energy and carbon of NLP.}
The energy cost of NLP training was surfaced in \cite{strubell2019energy}; subsequent work extended this to inference
\cite{lottick2019energy,bannour2021evaluating}.
Patterson et al.\ \cite{patterson2021carbon} quantified the carbon of large
model training and highlighted the role of data-centre location and grid mix.
For inference, Samsi et al.\ \cite{samsi2023words} measured the energy of LLM
inference on GPUs, and Luccioni et al.\ \cite{luccioni2024power} showed that
generative, multi-purpose models consume substantially more energy per
query than task-specific ones.
EcoLogits \cite{ecologits2024} provides per-token electricity estimates for
major API models, which we build on here; energy scaling laws have been
proposed to predict GPU energy for generative inference from its compute
requirements \cite{iyengar2025diffusion}, and profiles how input-token energy differs from
output-token energy \cite{ruf2026context}.
These works characterise single inference calls; we study how cost compounds
across the multi-call loops of agentic systems.

\paragraph{Cost-aware agents.}
Agent evaluation has increasingly called for reporting cost alongside accuracy
\cite{kapoor2025agents}, and cascading strategies such as FrugalGPT
\cite{chen2024frugalgpt} trade model size against call volume to reduce API
spend.
Separately, Huang et al.\ \cite{huang2023selfcorrect} show that intrinsic
self-correction without external feedback does not reliably improve reasoning,
which motivates measuring when additional refinement steps stop paying off.
We extend this cost-aware view from monetary cost to energy and carbon.

\section{Carbon Estimation}
\label{sec:carbon}

Detailed carbon data for cloud AI inference services is generally not
publicly disclosed by API providers \cite{hankendi2025transparency}.
We therefore estimate inference carbon using a methodology that derives per-output-token electricity factors from GPU and server power
models calibrated to known hardware configurations \cite{ecologits2024}.
For a run with $T$ completion tokens, model-specific electricity factor $e$
(kWh/token), grid factor $g$ (kgCO$_2$e/kWh), and scenario-embedded factor $b$
(kgCO$_2$e/token), we report
$$
\mathrm{CO2e}_{\mathrm{total}} = T \cdot e \cdot g + T \cdot b.
$$
Thus, the energy columns in our tables report the electricity term only,
while the carbon columns report the sum of Scope~2 electricity emissions and
the scenario's embedded per-token component. In both studies, the estimator is
applied to completion tokens only.
Because regional capacity for the required models differed at the time of the
experiments, the two studies were run in different cloud regions (Ireland for
Study~1, Sweden for Study~2), and each study applies the grid factor of its
region. Absolute carbon values are therefore not comparable across studies;
Section~\ref{sec:crossstudy} provides a like-for-like comparison under a
common European grid factor.
To enable reproducibility and transparency, we use an open carbon footprint data model \cite{carbontool2024}, which encodes per-configuration token totals
into shareable benchmark URLs for cross-configuration comparison.

\section{Study 1: Iterative Self-Refinement}
\label{sec:study1}

\subsection{Setup}

\paragraph{Benchmark.}
We evaluate on the MATH dataset \cite{hendrycks2021math}, specifically
Levels~3--4 ($N=100$ problems) and Levels~4--5 ($N=100$ problems).
Gold answers are extracted from the final boxed answer in each solution;
predictions are matched via float comparison, LaTeX fraction parsing, and
normalised string comparison.

\paragraph{Configurations.}
We evaluate six configurations:
GPT-4.1-mini at $k \in \{1, 2, 3, 4, 5\}$ and GPT-4.1 at $k=1$ as the
quality baseline.
All calls are made via Azure OpenAI using location \texttt{northeurope} at temperature~0.

\paragraph{Agent loop.}
The parameter $k$ is the total number of LLM calls made per problem.
Step~1: the model solves the problem (\emph{solve}).
Step~2: the model critiques its own solution (\emph{critique}).
Step~3+: the model verifies and revises (\emph{revise}).
Each call receives the problem together with all previous responses, and the
answer extracted from the final ($k$-th) response is scored. Thus $k=1$ is a
single-pass solve, $k=2$ adds one critique, and each further step adds one
verify-and-revise round. The depth is forced: every task makes exactly $k$
API calls, with no early stopping even if the model declares its answer
correct.

\paragraph{Carbon estimation.}
Using Microsoft's Azure regions list \cite{azure_regions_list}, we map \texttt{northeurope} to Ireland and apply its 2022 grid factor of 0.3476~kgCO$_2$e/kWh \cite{iea_emissions_factors_2022}. 
For GPT-4.1-mini, we use an electricity coefficient per output token of \href{https://borisruf.github.io/carbon-footprint-modeling-tool/show.html?scenario_id=gpt41mini-ecologits2-token&show_consumption=true}{0.4800~mWh} (modelled as a dense architecture); for GPT-4.1, we use \href{https://borisruf.github.io/carbon-footprint-modeling-tool/show.html?scenario_id=gpt41-ecologits2-token&show_consumption=true}{2.1857~mWh} (modelled as a mixture-of-experts architecture), following the methodology in Section~\ref{sec:carbon}. In Table~\ref{tab:math}, \emph{Wh/task} is the electricity term only; \emph{mgCO$_2$e} adds the scenario's embedded per-token component.

\subsection{Results}

\begin{figure*}[!t]
  \centering
  \begin{subfigure}[t]{0.32\linewidth}
    \includegraphics[width=\linewidth]{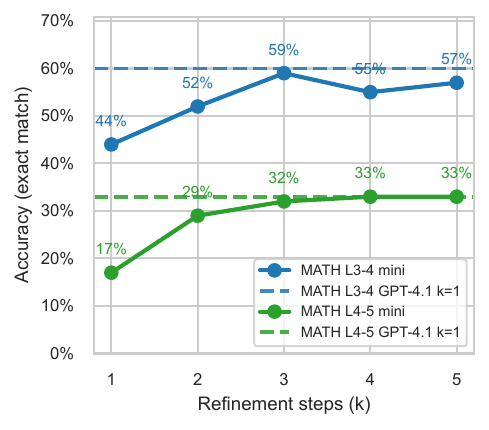}
    \caption{Accuracy vs.\ $k$}
  \end{subfigure}
  \hfill
  \begin{subfigure}[t]{0.32\linewidth}
    \includegraphics[width=\linewidth]{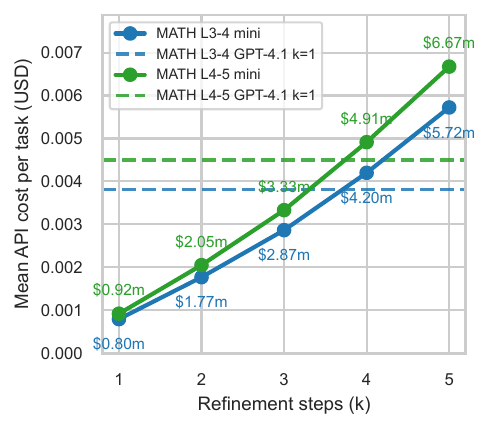}
    \caption{API cost vs.\ $k$}
  \end{subfigure}
  \hfill
  \begin{subfigure}[t]{0.32\linewidth}
    \includegraphics[width=\linewidth]{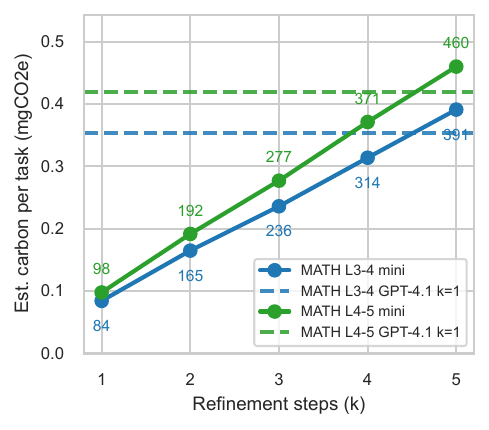}
    \caption{Est.\ carbon vs.\ $k$}
  \end{subfigure}
  \caption{Study~1: accuracy, cost, and carbon vs.\ $k$ on MATH Level~3--4 and
           Level~4--5. Dashed line: GPT-4.1 $k=1$.}
  \label{fig:study1}
\end{figure*}

Table~\ref{tab:math} and Figure~\ref{fig:study1} show
accuracy, cost, and carbon for each configuration on both MATH splits.

\begin{table*}[!t]
\centering
\footnotesize
\setlength{\tabcolsep}{1.8pt}
\caption{Mean per-task metrics on MATH ($N=100$ per split) using
         estimated per-output-token carbon scenarios and 0.3476~kgCO$_2$e/kWh.
         Accuracy brackets give 95\% Wilson confidence intervals.}
\label{tab:math}
\begin{subtable}[t]{0.49\textwidth}
  \centering
  \caption{\strut Level~3--4}
  \label{tab:main}
  \begin{tabular}{@{}lrrrrr@{}}
  \toprule
  Model & $k$ & Accuracy & Cost/task & Wh/task$^\dagger$ & mgCO$_2$e$^\dagger$ \\
  \midrule
  GPT-4.1-mini & 1 & 44\% {\scriptsize[35,54]} & .00080 &  0.22 & 84 \\
  GPT-4.1-mini & 2 & 52\% {\scriptsize[42,62]} & .00177 & 0.44 & 165 \\
  GPT-4.1-mini & 3 & \textbf{59\%} {\scriptsize[49,68]} & .00287 & 0.62 & 236 \\
  GPT-4.1-mini & 4 & 55\% {\scriptsize[45,64]} & .00420 & 0.83 & 314 \\
  GPT-4.1-mini & 5 & 57\% {\scriptsize[47,66]} & .00572 & 1.04 & 391 \\
  \midrule
  GPT-4.1      & 1 & \textbf{60\%} {\scriptsize[50,69]} & .00381 &  0.97 & 354 \\
  \bottomrule
  \end{tabular}
\end{subtable}\hfill
\begin{subtable}[t]{0.49\textwidth}
  \centering
  \caption{\strut Level~4--5}
  \label{tab:l45}
  \begin{tabular}{@{}lrrrrr@{}}
  \toprule
  Model & $k$ & Accuracy & Cost/task & Wh/task$^\dagger$ & mgCO$_2$e$^\dagger$ \\
  \midrule
  GPT-4.1-mini & 1 & 17\% {\scriptsize[11,26]} & .00092 &  0.26 & 98 \\
  GPT-4.1-mini & 2 & 29\% {\scriptsize[21,39]} & .00205 & 0.51 & 192 \\
  GPT-4.1-mini & 3 & \textbf{32\%} {\scriptsize[24,42]} & .00333 & 0.73 & 277 \\
  GPT-4.1-mini & 4 & \textbf{33\%} {\scriptsize[25,43]} & .00491 & 0.98 & 371 \\
  GPT-4.1-mini & 5 & \textbf{33\%} {\scriptsize[25,43]} & .00667 & 1.22 & 460 \\
  \midrule
  GPT-4.1      & 1 & \textbf{33\%} {\scriptsize[25,43]} & .00450 &  1.15 & 419 \\
  \bottomrule
  \end{tabular}
\end{subtable}
\smallskip\\
{\footnotesize $^\dagger$~Estimated from carbon scenarios; \emph{Wh/task} is electricity only, while \emph{mgCO$_2$e} combines electricity and the scenario's embedded per-token component.}
\end{table*}

\paragraph{Self-refinement closes the accuracy gap.}
On Level~3--4, GPT-4.1-mini at $k=1$ achieves 44\% accuracy, 16 percentage
points (pp) below GPT-4.1 (60\%).
By $k=3$, mini reaches 59\%, closing the gap to within 1~pp at 25\% lower
API cost.
The same pattern holds on Level~4--5: mini at $k=1$ scores 17\% vs.\ 33\%
for GPT-4.1, again exactly 16~pp behind, and $k=3$ closes to 32\% ($-$1~pp)
at 26\% lower cost.
Paired McNemar tests on the same 100 problems confirm that the gain from
$k=1$ to $k=3$ is significant on both splits ($p=0.001$), while the remaining
difference between mini $k=3$ and GPT-4.1 is not ($p=1.0$ on both splits).

\paragraph{Accuracy plateaus beyond $k=3$.}
Additional refinement steps yield no significant improvement.
On Level~3--4, accuracy at $k=4$ and $k=5$ (55\% and 57\%) is lower than at
$k=3$, but this difference lies within sampling noise (McNemar $p=0.50$ for
$k=3$ vs.\ $k=4$; 12 problems lost vs.\ 8 gained).
On Level~4--5, accuracy stays at 33\% from $k=4$ onward, matching GPT-4.1.

\paragraph{Carbon rises monotonically with refinement depth.}
On both splits, the most accurate non-baseline refinement setting ($k=3$)
emits approximately $2.8\times$ more carbon than the $k=1$ mini baseline.
Under the current per-output-token carbon scenario, however, mini $k=3$
remains below the GPT-4.1 single-pass baseline on both splits: 236 vs.\ 354~mgCO$_2$e
on Level~3--4 and 277 vs.\ 419~mgCO$_2$e on Level~4--5.
Carbon continues to increase with additional refinement steps and exceeds the
GPT-4.1 baseline only at $k=5$ on both splits.
The \emph{carbon--accuracy crossover} (beyond which further steps add carbon
without a measurable accuracy gain) is at $k=3$ on both splits; the observed
peak is $k=3$ on Level~3--4 and $k=4$ on Level~4--5, but the two are not
distinguishable at $N=100$.
The \emph{cost--accuracy crossover} (where refinement exceeds the upgrade
cost) is $k=4$ on both splits.
These thresholds are distinct; attending to both is essential for sustainable
deployment decisions.

\paragraph{Reproducibility (Carbon Estimates).}
Carbon comparisons based on token totals and the Irish grid factor are reproducible via the carbon footprint modelling tool for both MATH splits:
\href{https://borisruf.github.io/carbon-footprint-modeling-tool/benchmark.html?title=Self-Refinement+Study+MATH+L3-4&ids[]=gpt41mini-ecologits2-token&factors[]=46422&labels[]=mini+k=1&source_ids[]=ireland&ids[]=gpt41mini-ecologits2-token&factors[]=90922&labels[]=mini+k=2&source_ids[]=ireland&ids[]=gpt41mini-ecologits2-token&factors[]=130135&labels[]=mini+k=3&source_ids[]=ireland&ids[]=gpt41mini-ecologits2-token&factors[]=173103&labels[]=mini+k=4&source_ids[]=ireland&ids[]=gpt41mini-ecologits2-token&factors[]=215710&labels[]=mini+k=5&source_ids[]=ireland&ids[]=gpt41-ecologits2-token&factors[]=44335&labels[]=GPT-4.1+k=1&source_ids[]=ireland}{Level~3--4}
and
\href{https://borisruf.github.io/carbon-footprint-modeling-tool/benchmark.html?title=Self-Refinement+Study+MATH+L4-5&ids[]=gpt41mini-ecologits2-token&factors[]=53879&labels[]=mini+k=1&source_ids[]=ireland&ids[]=gpt41mini-ecologits2-token&factors[]=105667&labels[]=mini+k=2&source_ids[]=ireland&ids[]=gpt41mini-ecologits2-token&factors[]=152805&labels[]=mini+k=3&source_ids[]=ireland&ids[]=gpt41mini-ecologits2-token&factors[]=204756&labels[]=mini+k=4&source_ids[]=ireland&ids[]=gpt41mini-ecologits2-token&factors[]=253760&labels[]=mini+k=5&source_ids[]=ireland&ids[]=gpt41-ecologits2-token&factors[]=52531&labels[]=GPT-4.1+k=1&source_ids[]=ireland}{Level~4--5}
.

\section{Study 2: Tool-Augmented Agents}
\label{sec:study2}

\subsection{Setup}

\paragraph{Benchmark.}
SWE-bench Lite \cite{jimenez2024swebench} is a curated subset of 300 real
GitHub issues drawn from popular Python repositories (Django, Matplotlib,
SymPy, scikit-learn, and others).
Each task requires the agent to produce a patch that makes a set of
previously-failing tests pass.
Success is measured as pass@1: whether the generated patch resolves the issue
in a single attempt as assessed by the official evaluation harness.
Each model was run on the first 50 instances of the SWE-bench Lite test split
(in dataset order: 6 Astropy and 44 Django issues), with up to three attempts
per instance. Two instances failed to complete for every model and are
excluded, leaving 48 for GPT-4.1-mini and GPT-5.3-codex. For Claude Opus~4.6,
three further instances timed out and are excluded, leaving 45. On the 45 instances common
to all three models, resolution rates are 26.7\% (GPT-4.1-mini), 68.9\%
(Claude Opus~4.6), and 82.2\% (GPT-5.3-codex); counting the three timeouts
as unresolved instead gives 64.6\% for Claude Opus~4.6. The ranking is
therefore unaffected by how the timed-out instances are handled.

\paragraph{Agent framework.}
We use the OpenHands agent \cite{openhands2024} with the default tool preset:
bash execution, file editing, and a structured task instruction.
Each task runs in a dedicated Docker container pre-built for the target
repository commit, providing full process isolation and a reproducible
execution environment.
The agent loop continues for up to 500 LLM iterations; a critic module
checks for task completion at the end of each attempt.

\paragraph{Configurations.}
We compare three models: GPT-4.1-mini (the same cheap baseline as Study~1),
Claude Opus~4.6 (Anthropic's flagship model), and GPT-5.3-codex (a
state-of-the-art reasoning-optimised model).
All calls are routed through a LiteLLM proxy to Azure OpenAI or the
Anthropic API (Sweden Central region) at each model's default temperature.

\paragraph{Carbon estimation.}
We use Sweden's 2022 grid-emissions factor, 0.0256~kgCO$_2$e/kWh \cite{iea_emissions_factors_2022}.
We use electricity coefficients per output token of \href{https://borisruf.github.io/carbon-footprint-modeling-tool/show.html?scenario_id=gpt41mini-ecologits2-token&show_consumption=true}{0.4800~mWh} for GPT-4.1-mini, \href{https://borisruf.github.io/carbon-footprint-modeling-tool/show.html?scenario_id=claudeopus46-ecologits2-token&show_consumption=true}{6.6824~mWh} for Claude Opus~4.6, and \href{https://borisruf.github.io/carbon-footprint-modeling-tool/show.html?scenario_id=gpt53codex-ecologits2-token&show_consumption=true}{2.4322~mWh} for GPT-5.3-codex, again using the Section~\ref{sec:carbon} methodology. As in Study~1, the reported carbon values add the scenarios' embedded per-token component to the electricity-based estimate. These values cover LLM inference only; they exclude container startup, repository installation, tool execution, and test-suite compute inside the Docker environment.

\subsection{Results}

\begin{figure*}[!t]
  \centering
  \begin{subfigure}[t]{0.32\linewidth}
    \includegraphics[width=\linewidth]{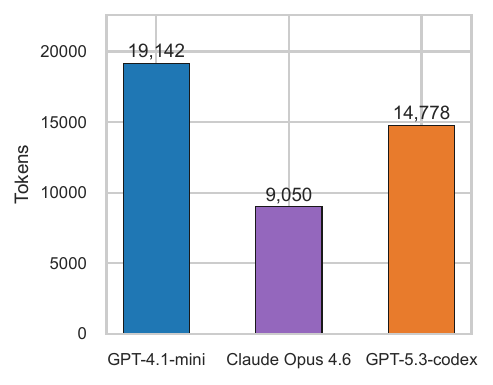}
    \caption{Output tokens per resolved task}
  \end{subfigure}\hfill
  \begin{subfigure}[t]{0.32\linewidth}
    \includegraphics[width=\linewidth]{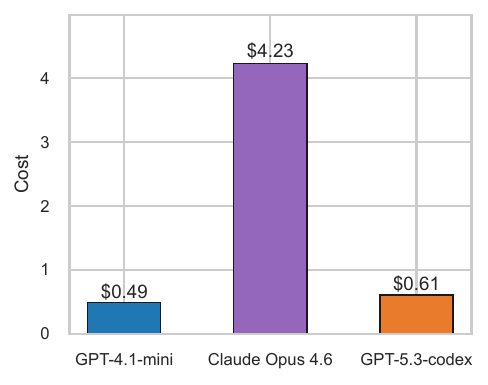}
    \caption{Mean API cost per resolved task}
  \end{subfigure}\hfill
  \begin{subfigure}[t]{0.32\linewidth}
    \includegraphics[width=\linewidth]{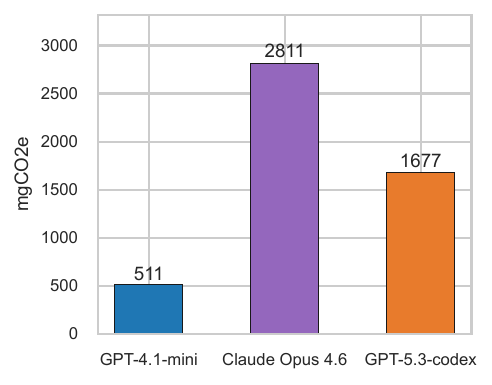}
    \caption{Est.\ carbon per resolved task}
  \end{subfigure}
  \caption{Study~2: output tokens, API cost, and estimated carbon per resolved task on SWE-bench Lite.}
  \label{fig:study2}
\end{figure*}

Table~\ref{tab:swebench} and Figure~\ref{fig:study2} summarise the results
for all three models.

\begin{table*}[!b]
\centering
\small
\setlength{\tabcolsep}{4pt}
\caption{Core per-task metrics on SWE-bench Lite (iterations retained) using
         Sweden's 2022 grid factor. Resolution brackets give 95\% Wilson
         confidence intervals.}
\label{tab:swebench}
\begin{tabular}{lrrrrrr}
\toprule
Model & $N$ & Iterations & Cost/task & kWh/task$^\dagger$ & mgCO$_2$e/task$^\dagger$ & Resolution (pass@1) \\
\midrule
GPT-4.1-mini    & 48 & 54.4 & 0.123 & 0.002297 & 128 & 25.0\% {\scriptsize[15,39]} \\
Claude Opus~4.6 & 45 & \textbf{23.7} & 2.912 & 0.041661 & 1{,}936 & 68.9\% {\scriptsize[54,80]} \\
GPT-5.3-codex   & 48 & 39.4 & 0.495 & 0.029204 & 1{,}363 & \textbf{81.2\%} {\scriptsize[68,90]} \\
\bottomrule
\end{tabular}
\smallskip\\
{\footnotesize $^\dagger$~Estimated from carbon scenarios; \emph{kWh/task} is electricity only, while \emph{mgCO$_2$e/task} combines electricity and the scenario's embedded per-token component.}
\end{table*}

\paragraph{Resolution rate.}
GPT-5.3-codex leads with 81.2\% pass@1, followed by Claude Opus~4.6 at
68.9\% and GPT-4.1-mini at 25.0\%.
On the 45 shared instances, GPT-5.3-codex resolves six tasks that Claude
Opus~4.6 does not and none in reverse (McNemar $p=0.031$); given the sample
size, this ordering should nevertheless be read as indicative.
Claude Opus~4.6 converges in just 23.7 iterations on average---the fewest
of the three---while GPT-4.1-mini requires 54.4.

\paragraph{Cost and carbon per task.}
Claude Opus~4.6 is the most expensive per task (\$2.91), roughly 6$\times$
GPT-5.3-codex (\$0.495) and 24$\times$ GPT-4.1-mini (\$0.123).
At the task level, GPT-4.1-mini is the lowest-energy and lowest-carbon model
(0.0023~kWh, 128~mgCO$_2$e), followed by GPT-5.3-codex
(0.0292~kWh, 1{,}363~mgCO$_2$e), then Claude Opus~4.6
(0.0417~kWh, 1{,}936~mgCO$_2$e).

\paragraph{Carbon per resolved task.}
Normalised by outcome---carbon per \emph{resolved} task---the ordering is
unchanged but the gaps narrow: GPT-4.1-mini emits 511~mgCO$_2$e per resolved task,
GPT-5.3-codex emits 1{,}677~mgCO$_2$e (3.3$\times$ mini, down from
10.6$\times$ per task), and
Claude Opus~4.6 emits 2{,}811~mgCO$_2$e (5.5$\times$ mini, down from
15.1$\times$ per task).
Thus, GPT-5.3-codex leads on pass@1, while GPT-4.1-mini is most carbon-efficient
per resolved task under this token-based estimate.

\paragraph{Reproducibility (Carbon Estimates).}
Using the carbon footprint modelling tool, we reproduce \href{https://borisruf.github.io/carbon-footprint-modeling-tool/benchmark.html?title=SWE-bench+Lite+Study&ids[]=claudeopus46-ecologits2-token&factors[]=280546&labels[]=Claude+Opus+4.6&source_ids[]=sweden&ids[]=gpt41mini-ecologits2-token&factors[]=229699&labels[]=GPT-4.1-mini&source_ids[]=sweden&ids[]=gpt53codex-ecologits2-token&factors[]=576330&labels[]=GPT-5.3-codex&source_ids[]=sweden}{carbon comparisons} across configurations.

\section{Discussion}

\paragraph{Study~1: when is self-refinement worth it?}
Across both difficulty levels, self-refinement with GPT-4.1-mini is
\emph{cost-competitive} with a GPT-4.1 upgrade at $k=3$ ($\approx$25\%
cheaper, $-$1pp accuracy), and under the current token-based carbon scenario
it also remains below the GPT-4.1 single-pass baseline at $k=3$ on both
MATH splits. Carbon nevertheless rises monotonically with refinement depth,
so the sustainability gain relative to GPT-4.1 is scenario-dependent and
does not hold once refinement is pushed further.

\paragraph{Study~2: capability and carbon efficiency diverge.}
Study~2 reveals a different pattern.
In the tool-augmented, grounded setting of SWE-bench Lite, the highest-pass@1
model in this sample (GPT-5.3-codex) is substantially more accurate and converges faster
than GPT-4.1-mini, but it is not dominant on every sustainability axis:
GPT-4.1-mini remains lower on estimated cost and carbon per resolved task,
while Claude Opus~4.6 converges in the fewest iterations.
Under output-token accounting, capability therefore buys outcomes that the
cheaper model cannot deliver, but not lower carbon per outcome.
Grounded tool use does penalise weak models: GPT-4.1-mini fails more often and
iterates longer, and so accumulates far more prompt context per task (see
below). Whether this narrows or reverses the per-resolved-task gap under
prompt-inclusive accounting remains open.

\paragraph{Diminishing returns in Study~1.}
On neither split does refinement beyond $k=3$ yield a significant accuracy
gain, while each further step adds carbon at a roughly constant rate.
The lower accuracy at $k=4$ and $k=5$ on Level~3--4 is within sampling noise.
Prior work finds that intrinsic self-correction can turn correct answers into
incorrect ones \cite{huang2023selfcorrect}; whether this effect becomes
significant with larger samples is left for future work.

\paragraph{Cross-study comparisons.}
\label{sec:crossstudy}
Owing to model availability constraints, Study~1 ran in a data center in Ireland (Azure North Europe, $\approx 348$~gCO$_2$e/kWh) and Study~2 in Sweden ($\approx 25.6$~gCO$_2$e/kWh)\cite{iea_emissions_factors_2022}, so their
reported carbon figures are not directly comparable.
For a like-for-like comparison, we use the carbon footprint modelling tool to create a
\href{https://borisruf.github.io/carbon-footprint-modeling-tool/benchmark.html?title=Cross-Study+Carbon+Comparison+(average%2C+Europe+2022)&ids[]=gpt41mini-ecologits2-token&factors[]=46422&labels[]=S1+mini+k%3D1&source_ids[]=europe&ids[]=gpt41mini-ecologits2-token&factors[]=90922&labels[]=S1+mini+k%3D2&source_ids[]=europe&ids[]=gpt41mini-ecologits2-token&factors[]=130135&labels[]=S1+mini+k%3D3&source_ids[]=europe&ids[]=gpt41mini-ecologits2-token&factors[]=173103&labels[]=S1+mini+k%3D4&source_ids[]=europe&ids[]=gpt41mini-ecologits2-token&factors[]=215710&labels[]=S1+mini+k%3D5&source_ids[]=europe&ids[]=gpt41-ecologits2-token&factors[]=44335&labels[]=S1+GPT-4.1+k%3D1&source_ids[]=europe&ids[]=gpt41mini-ecologits2-token&factors[]=229699&labels[]=S2+GPT-4.1-mini&source_ids[]=europe&ids[]=gpt53codex-ecologits2-token&factors[]=576330&labels[]=S2+GPT-5.3-codex&source_ids[]=europe&ids[]=claudeopus46-ecologits2-token&factors[]=280546&labels[]=S2+Claude+Opus+4.6&source_ids[]=europe}{carbon benchmark} that
applies Europe's 2022 average grid-emissions factor ($\approx 248$~gCO$_2$e/kWh) to every configuration, enabling direct comparison across both studies. For Study~1, we use the Level~3--4 split as representative; the Level~4--5 split shows the same ordering.

\paragraph{Token composition and the limits of output-token costing.}
Figure~\ref{fig:tokens} (Appendix) quantifies a structural gap in output-token-based carbon
accounting across both studies.
In Study~1 (Figure~\ref{fig:tokens}\,a), prompt tokens rise from 22\% of all tokens at $k=1$
to 73\% at $k=5$ as prior responses are prepended to the context. Because
our current carbon scenario charges only for completion tokens, it captures a shrinking
share of token processing as $k$ increases.
In Study~2 (Figure~\ref{fig:tokens}\,b), the imbalance is stronger: prompt tokens make up
over 99\% of all tokens at every iteration across all three models, growing from a few
thousand at iteration~1 to tens of thousands by convergence. Weaker models therefore
accumulate more context-processing work in addition to making more calls, which
output-only accounting does not capture. Recent direct measurements on open
models further show that input tokens have far lower energy intensity than
output tokens, even in long-context settings, although the size of the gap
depends on the serving stack \cite{ruf2026context}. Prompt-token dominance
by count therefore does not imply prompt-token dominance in carbon. The main
limitation of our estimator is that it can understate absolute footprint and
misstate comparisons when prompt growth differs sharply across agents, rather
than that token-count shares map directly to carbon shares.

\paragraph{Limitations.}
Study~1 evaluates a single model family on a mathematical reasoning benchmark,
and Study~2 is limited to 45--48 SWE-bench Lite instances per model, drawn
from only two repositories (Astropy and Django), so the
results may not generalise to other models, repositories, agent frameworks,
or deployment environments.
With $N=100$ per split, accuracy estimates in Study~1 carry 95\% confidence
intervals of roughly $\pm$10~pp, so only large differences are resolvable.
Our carbon estimates treat model inference as the sole emission source. In
particular, they include the per-token embedded component supplied by the selected carbon
scenarios, but they do not include prompt-token processing, Docker container
execution, test runs, networking, or additional data-centre overheads such as
cooling beyond what is already implicit in the scenario assumptions.
Direct energy measurements on open models would provide a stronger basis for comparison.

\section{Conclusion}

We have presented two empirical studies that quantify the energy and carbon
footprint of agentic AI across settings of increasing complexity.
In purely verbal self-refinement (Study~1), iterative refinement with a
cheap model can close a 16-percentage-point accuracy gap at lower API cost
but always at a nearly tripled carbon footprint relative to the mini baseline,
while remaining below the GPT-4.1 single-pass baseline under the current
token-based carbon scenario across two difficulty levels of MATH.
In tool-augmented agentic coding (Study~2), GPT-5.3-codex achieves the highest
pass@1 in this sample and
resolves 3.2$\times$ more SWE-bench Lite tasks than GPT-4.1-mini, but at
approximately 3.3$\times$ higher estimated carbon per resolved task under
output-token accounting, which leaves out the prompt tokens that make up over
99\% of tokens in this setting, as well as tool execution. This highlights that capability gains and
sustainability gains need not move in the same direction.

Together, the two studies suggest that the carbon cost of agentic AI depends
on task structure: in verbal self-refinement, cheaper models with more steps
accumulate carbon with diminishing accuracy returns, whereas in grounded tool
use, stronger models deliver outcomes that cheaper models cannot, at higher
estimated carbon per outcome. Extending the accounting to prompt tokens and
tool execution is the main next step to determine whether this ordering
holds for the full agentic footprint.

\bibliography{references}

\begin{thebibliography}{25}
\expandafter\ifx\csname natexlab\endcsname\relax\def\natexlab#1{#1}\fi
\providecommand{\url}[1]{\texttt{#1}}
\providecommand{\href}[2]{#2}
\providecommand{\path}[1]{#1}
\providecommand{\DOIprefix}{doi:}
\providecommand{\ArXivprefix}{arXiv:}
\providecommand{\URLprefix}{URL: }
\providecommand{\Pubmedprefix}{pmid:}
\providecommand{\doi}[1]{\href{http://dx.doi.org/#1}{\path{#1}}}
\providecommand{\Pubmed}[1]{\href{pmid:#1}{\path{#1}}}
\providecommand{\bibinfo}[2]{#2}
\ifx\xfnm\relax \def\xfnm[#1]{\unskip,\space#1}\fi
\bibitem[{Madaan et~al.(2023)Madaan, Tandon, Gupta, Hallinan, Gao, Wiegreffe, Alon, Dziri, Prabhumoye, Yang, Gupta, Majumder, Hermann, Welleck, Yazdanbakhsh, and Clark}]{madaan2023selfrefine}
\bibinfo{author}{A.~Madaan}, \bibinfo{author}{N.~Tandon}, \bibinfo{author}{P.~Gupta}, \bibinfo{author}{S.~Hallinan}, \bibinfo{author}{L.~Gao}, \bibinfo{author}{S.~Wiegreffe}, \bibinfo{author}{U.~Alon}, \bibinfo{author}{N.~Dziri}, \bibinfo{author}{S.~Prabhumoye}, \bibinfo{author}{Y.~Yang}, \bibinfo{author}{S.~Gupta}, \bibinfo{author}{B.~P. Majumder}, \bibinfo{author}{K.~Hermann}, \bibinfo{author}{S.~Welleck}, \bibinfo{author}{A.~Yazdanbakhsh}, \bibinfo{author}{P.~Clark},
\newblock \bibinfo{title}{Self-refine: iterative refinement with self-feedback},
\newblock in: \bibinfo{booktitle}{Proceedings of the 37th International Conference on Neural Information Processing Systems}, NIPS, \bibinfo{publisher}{Curran Associates Inc.}, \bibinfo{year}{2023}.
\bibitem[{Shinn et~al.(2023)Shinn, Cassano, Gopinath, Narasimhan, and Yao}]{shinn2023reflexion}
\bibinfo{author}{N.~Shinn}, \bibinfo{author}{F.~Cassano}, \bibinfo{author}{A.~Gopinath}, \bibinfo{author}{K.~Narasimhan}, \bibinfo{author}{S.~Yao},
\newblock \bibinfo{title}{Reflexion: Language agents with verbal reinforcement learning},
\newblock in: \bibinfo{booktitle}{Neural Information Processing Systems}, \bibinfo{publisher}{Neural Information Processing Systems Foundation, Inc. (NeurIPS)}, \bibinfo{year}{2023}, pp. \bibinfo{pages}{8634--8652}.
\bibitem[{Yao et~al.(2023)Yao, Zhao, Yu, Du, Shafran, Narasimhan, and Cao}]{yao2023react}
\bibinfo{author}{S.~Yao}, \bibinfo{author}{J.~Zhao}, \bibinfo{author}{D.~Yu}, \bibinfo{author}{N.~Du}, \bibinfo{author}{I.~Shafran}, \bibinfo{author}{K.~R. Narasimhan}, \bibinfo{author}{Y.~Cao},
\newblock \bibinfo{title}{React: Synergizing reasoning and acting in language models},
\newblock in: \bibinfo{booktitle}{The 11th International Conference on Learning Representations}, \bibinfo{year}{2023}.
\bibitem[{Wang et~al.(2025)Wang, Li, Song, Xu, Tang, Zhuge, Pan, Song, Li, Singh, Tran, Li, Ma, Zheng, Qian, Shao, Muennighoff, Zhang, Hui, Lin, Brennan, Peng, Ji, and Neubig}]{openhands2024}
\bibinfo{author}{X.~Wang}, \bibinfo{author}{B.~Li}, \bibinfo{author}{Y.~Song}, \bibinfo{author}{F.~F. Xu}, \bibinfo{author}{X.~Tang}, \bibinfo{author}{M.~Zhuge}, \bibinfo{author}{J.~Pan}, \bibinfo{author}{Y.~Song}, \bibinfo{author}{B.~Li}, \bibinfo{author}{J.~Singh}, \bibinfo{author}{H.~H. Tran}, \bibinfo{author}{F.~Li}, \bibinfo{author}{R.~Ma}, \bibinfo{author}{M.~Zheng}, \bibinfo{author}{B.~Qian}, \bibinfo{author}{Y.~Shao}, \bibinfo{author}{N.~Muennighoff}, \bibinfo{author}{Y.~Zhang}, \bibinfo{author}{B.~Hui}, \bibinfo{author}{J.~Lin}, \bibinfo{author}{R.~Brennan}, \bibinfo{author}{H.~Peng}, \bibinfo{author}{H.~Ji}, \bibinfo{author}{G.~Neubig},
\newblock \bibinfo{title}{Open{H}ands: An open platform for {AI} software developers as generalist agents},
\newblock in: \bibinfo{booktitle}{The 13th International Conference on Learning Representations}, \bibinfo{year}{2025}.
\bibitem[{Chavan et~al.(2024)Chavan, Magazine, Kushwaha, Debbah, and Gupta}]{chavan2024faster}
\bibinfo{author}{A.~Chavan}, \bibinfo{author}{R.~Magazine}, \bibinfo{author}{S.~Kushwaha}, \bibinfo{author}{M.~Debbah}, \bibinfo{author}{D.~Gupta},
\newblock \bibinfo{title}{Faster and lighter {LLMs}: a survey on current challenges and way forward},
\newblock in: \bibinfo{booktitle}{Proceedings of the 33rd International Joint Conference on Artificial Intelligence}, IJCAI, \bibinfo{year}{2024}.
\bibitem[{Jimenez et~al.(2024)Jimenez, Yang, Wettig, Yao, Pei, Press, and Narasimhan}]{jimenez2024swebench}
\bibinfo{author}{C.~E. Jimenez}, \bibinfo{author}{J.~Yang}, \bibinfo{author}{A.~Wettig}, \bibinfo{author}{S.~Yao}, \bibinfo{author}{K.~Pei}, \bibinfo{author}{O.~Press}, \bibinfo{author}{K.~Narasimhan},
\newblock \bibinfo{title}{{SWE}-bench: Can language models resolve real-world {GitHub} issues?},
\newblock in: \bibinfo{booktitle}{International Conference on Learning Representations}, \bibinfo{year}{2024}.
\bibitem[{Jiang et~al.(2024)Jiang, Peng, Feng, Li, and Li}]{jiang2024llms}
\bibinfo{author}{Z.~Jiang}, \bibinfo{author}{H.~Peng}, \bibinfo{author}{S.~Feng}, \bibinfo{author}{F.~Li}, \bibinfo{author}{D.~Li},
\newblock \bibinfo{title}{{LLMs} can find mathematical reasoning mistakes by pedagogical chain-of-thought},
\newblock in: \bibinfo{booktitle}{Proceedings of the 33rd International Joint Conference on Artificial Intelligence}, IJCAI, \bibinfo{year}{2024}.
\bibitem[{Chen et~al.(2024)Chen, Dong, Shu, Zhang, Sesay, Karlsson, Fu, and Shi}]{chen2024autoagents}
\bibinfo{author}{G.~Chen}, \bibinfo{author}{S.~Dong}, \bibinfo{author}{Y.~Shu}, \bibinfo{author}{G.~Zhang}, \bibinfo{author}{J.~Sesay}, \bibinfo{author}{B.~Karlsson}, \bibinfo{author}{J.~Fu}, \bibinfo{author}{Y.~Shi},
\newblock \bibinfo{title}{{AutoAgents}: a framework for automatic agent generation},
\newblock in: \bibinfo{booktitle}{Proceedings of the 33rd International Joint Conference on Artificial Intelligence}, IJCAI, \bibinfo{year}{2024}.
\bibitem[{Strubell et~al.(2019)Strubell, Ganesh, and McCallum}]{strubell2019energy}
\bibinfo{author}{E.~Strubell}, \bibinfo{author}{A.~Ganesh}, \bibinfo{author}{A.~McCallum},
\newblock \bibinfo{title}{Energy and policy considerations for deep learning in {NLP}},
\newblock in: \bibinfo{editor}{A.~Korhonen}, \bibinfo{editor}{D.~Traum}, \bibinfo{editor}{L.~M{\`a}rquez} (Eds.), \bibinfo{booktitle}{Proceedings of the 57th Annual Meeting of the Association for Computational Linguistics}, \bibinfo{publisher}{Association for Computational Linguistics}, \bibinfo{year}{2019}, pp. \bibinfo{pages}{3645--3650}.
\bibitem[{Lottick et~al.(2019)Lottick, Susai, Friedler, and Wilson}]{lottick2019energy}
\bibinfo{author}{K.~Lottick}, \bibinfo{author}{S.~Susai}, \bibinfo{author}{S.~A. Friedler}, \bibinfo{author}{J.~P. Wilson},
\newblock \bibinfo{title}{Energy usage reports: Environmental awareness as part of algorithmic accountability},
\newblock \bibinfo{journal}{Workshop on Tackling Climate Change with Machine Learning at NeurIPS}  (\bibinfo{year}{2019}).
\bibitem[{Bannour et~al.(2021)Bannour, Ghannay, Névéol, and Ligozat}]{bannour2021evaluating}
\bibinfo{author}{N.~Bannour}, \bibinfo{author}{S.~Ghannay}, \bibinfo{author}{A.~Névéol}, \bibinfo{author}{A.-L. Ligozat},
\newblock \bibinfo{title}{Evaluating the carbon footprint of {NLP} methods: A survey and analysis of existing tools},
\newblock \bibinfo{journal}{SUSTAINLP} \bibinfo{volume}{27} (\bibinfo{year}{2021}) \bibinfo{pages}{11--21}.
\bibitem[{Patterson et~al.(2021)Patterson, Gonzalez, Le, Liang, Munguia, Rothchild, So, Texier, and Dean}]{patterson2021carbon}
\bibinfo{author}{D.~Patterson}, \bibinfo{author}{J.~Gonzalez}, \bibinfo{author}{Q.~Le}, \bibinfo{author}{C.~Liang}, \bibinfo{author}{L.-M. Munguia}, \bibinfo{author}{D.~Rothchild}, \bibinfo{author}{D.~So}, \bibinfo{author}{M.~Texier}, \bibinfo{author}{J.~Dean},
\newblock \bibinfo{title}{Carbon emissions and large neural network training},
\newblock \bibinfo{journal}{arXiv preprint arXiv:2104.10350}  (\bibinfo{year}{2021}).
\bibitem[{Samsi et~al.(2023)Samsi, Zhao, McDonald, Li, Michaleas, Jones, Bergeron, Kepner, Tiwari, and Gadepally}]{samsi2023words}
\bibinfo{author}{S.~Samsi}, \bibinfo{author}{D.~Zhao}, \bibinfo{author}{J.~McDonald}, \bibinfo{author}{B.~Li}, \bibinfo{author}{A.~Michaleas}, \bibinfo{author}{M.~Jones}, \bibinfo{author}{W.~Bergeron}, \bibinfo{author}{J.~Kepner}, \bibinfo{author}{D.~Tiwari}, \bibinfo{author}{V.~Gadepally},
\newblock \bibinfo{title}{From words to watts: Benchmarking the energy costs of large language model inference},
\newblock in: \bibinfo{booktitle}{2023 IEEE High Performance Extreme Computing Conference (HPEC)}, \bibinfo{publisher}{IEEE}, \bibinfo{year}{2023}, pp. \bibinfo{pages}{1--9}. \DOIprefix\doi{10.1109/HPEC58863.2023.10363447}.
\bibitem[{Luccioni et~al.(2024)Luccioni, Jernite, and Strubell}]{luccioni2024power}
\bibinfo{author}{S.~Luccioni}, \bibinfo{author}{Y.~Jernite}, \bibinfo{author}{E.~Strubell},
\newblock \bibinfo{title}{Power hungry processing: Watts driving the cost of {AI} deployment?},
\newblock in: \bibinfo{booktitle}{The 2024 ACM Conference on Fairness, Accountability, and Transparency}, FAccT '24, \bibinfo{publisher}{ACM}, \bibinfo{year}{2024}, pp. \bibinfo{pages}{85--99}. \DOIprefix\doi{10.1145/3630106.3658542}.
\bibitem[{Rincé and Banse(2025)}]{ecologits2024}
\bibinfo{author}{S.~Rincé}, \bibinfo{author}{A.~Banse},
\newblock \bibinfo{title}{Ecologits: Evaluating the environmental impacts of generative ai},
\newblock \bibinfo{journal}{Journal of Open Source Software} \bibinfo{volume}{10} (\bibinfo{year}{2025}) \bibinfo{pages}{7471}. \bibinfo{note}{Methodology: \url{https://ecologits.ai/0.9/methodology/llm_inference/}}.
\bibitem[{Iyengar et~al.(2026)Iyengar, Han, Ruf, Grari, Detyniecki, and Ermon}]{iyengar2025diffusion}
\bibinfo{author}{A.~Iyengar}, \bibinfo{author}{J.~Han}, \bibinfo{author}{B.~Ruf}, \bibinfo{author}{V.~Grari}, \bibinfo{author}{M.~Detyniecki}, \bibinfo{author}{S.~Ermon},
\newblock \bibinfo{title}{Energy scaling laws for diffusion models: Quantifying compute in image generation},
\newblock in: \bibinfo{booktitle}{The 2026 ACM Conference on Fairness, Accountability, and Transparency}, \bibinfo{year}{2026}, pp. \bibinfo{pages}{8029--8050}.
\bibitem[{Ruf and Detyniecki(2026)}]{ruf2026context}
\bibinfo{author}{B.~Ruf}, \bibinfo{author}{M.~Detyniecki},
\newblock \bibinfo{title}{The cost of context: Profiling the energy footprint of input tokens in large language models},
\newblock \bibinfo{journal}{HotCarbon / SIGENERGY Energy Informatics Review}  (\bibinfo{year}{2026}). \URLprefix \url{https://hotcarbon.org/assets/2026/paper-17.pdf}.
\bibitem[{Kapoor et~al.(2025)Kapoor, Stroebl, Siegel, Nadgir, and Narayanan}]{kapoor2025agents}
\bibinfo{author}{S.~Kapoor}, \bibinfo{author}{B.~Stroebl}, \bibinfo{author}{Z.~S. Siegel}, \bibinfo{author}{N.~Nadgir}, \bibinfo{author}{A.~Narayanan},
\newblock \bibinfo{title}{{AI} agents that matter},
\newblock \bibinfo{journal}{Transactions on Machine Learning Research} \bibinfo{volume}{2025-May} (\bibinfo{year}{2025}).
\bibitem[{Chen et~al.(2024)Chen, Zaharia, and Zou}]{chen2024frugalgpt}
\bibinfo{author}{L.~Chen}, \bibinfo{author}{M.~Zaharia}, \bibinfo{author}{J.~Zou},
\newblock \bibinfo{title}{Frugal{GPT}: How to use large language models while reducing cost and improving performance},
\newblock \bibinfo{journal}{Transactions on Machine Learning Research}  (\bibinfo{year}{2024}). \URLprefix \url{https://openreview.net/forum?id=cSimKw5p6R}, \bibinfo{note}{featured Certification}.
\bibitem[{Huang et~al.(2024)Huang, Chen, Mishra, Zheng, Yu, Song, and Zhou}]{huang2023selfcorrect}
\bibinfo{author}{J.~Huang}, \bibinfo{author}{X.~Chen}, \bibinfo{author}{S.~Mishra}, \bibinfo{author}{H.~S. Zheng}, \bibinfo{author}{A.~Yu}, \bibinfo{author}{X.~Song}, \bibinfo{author}{D.~Zhou},
\newblock \bibinfo{title}{Large language models cannot self-correct reasoning yet},
\newblock in: \bibinfo{editor}{B.~Kim}, \bibinfo{editor}{Y.~Yue}, \bibinfo{editor}{S.~Chaudhuri}, \bibinfo{editor}{K.~Fragkiadaki}, \bibinfo{editor}{M.~Khan}, \bibinfo{editor}{Y.~Sun} (Eds.), \bibinfo{booktitle}{International Conference on Learning Representations}, \bibinfo{year}{2024}, pp. \bibinfo{pages}{32808--32824}. \URLprefix \url{https://proceedings.iclr.cc/paper_files/paper/2024/file/8b4add8b0aa8749d80a34ca5d941c355-Paper-Conference.pdf}.
\bibitem[{Hankendi et~al.(2025)Hankendi, Coskun, and Sovacool}]{hankendi2025transparency}
\bibinfo{author}{C.~Hankendi}, \bibinfo{author}{A.~K. Coskun}, \bibinfo{author}{B.~Sovacool},
\newblock \bibinfo{title}{Why transparency matters for sustainable data centers and carbon-neutral artificial intelligence ({AI})},
\newblock \bibinfo{journal}{iScience} \bibinfo{volume}{28} (\bibinfo{year}{2025}) \bibinfo{pages}{113705}.
\bibitem[{Ruf and Detyniecki(2023)}]{carbontool2024}
\bibinfo{author}{B.~Ruf}, \bibinfo{author}{M.~Detyniecki},
\newblock \bibinfo{title}{Open and linked data model for carbon footprint scenarios},
\newblock in: \bibinfo{booktitle}{International Conference on Advanced Infocomm Technology}, \bibinfo{publisher}{IEEE}, \bibinfo{year}{2023}, pp. \bibinfo{pages}{423--427}.
\bibitem[{Hendrycks et~al.(2021)Hendrycks, Burns, Kadavath, Arora, Basart, Tang, Song, and Steinhardt}]{hendrycks2021math}
\bibinfo{author}{D.~Hendrycks}, \bibinfo{author}{C.~Burns}, \bibinfo{author}{S.~Kadavath}, \bibinfo{author}{A.~Arora}, \bibinfo{author}{S.~Basart}, \bibinfo{author}{E.~Tang}, \bibinfo{author}{D.~Song}, \bibinfo{author}{J.~Steinhardt},
\newblock \bibinfo{title}{Measuring mathematical problem solving with the {MATH} dataset},
\newblock \bibinfo{journal}{NeurIPS Datasets and Benchmarks}  (\bibinfo{year}{2021}).
\bibitem[{{Microsoft}(2026)}]{azure_regions_list}
\bibinfo{author}{{Microsoft}}, \bibinfo{title}{Azure regions list}, \bibinfo{howpublished}{\url{https://learn.microsoft.com/en-us/azure/reliability/regions-list}}, \bibinfo{year}{2026}. \bibinfo{note}{Accessed: 2026-05-12}.
\bibitem[{{International Energy Agency (IEA)}(2022)}]{iea_emissions_factors_2022}
\bibinfo{author}{{International Energy Agency (IEA)}}, \bibinfo{title}{Emissions factors 2022}, \bibinfo{howpublished}{\url{https://www.iea.org/data-and-statistics/data-product/emissions-factors-2022}}, \bibinfo{year}{2022}. \bibinfo{note}{Accessed: 2026-05-12}.

\end{thebibliography}


\clearpage
\onecolumn
\appendix
\section*{Appendix}
\section{Token Composition}

\begin{figure}[h]
  \centering
  \begin{subfigure}[t]{0.48\linewidth}
    \includegraphics[width=\linewidth]{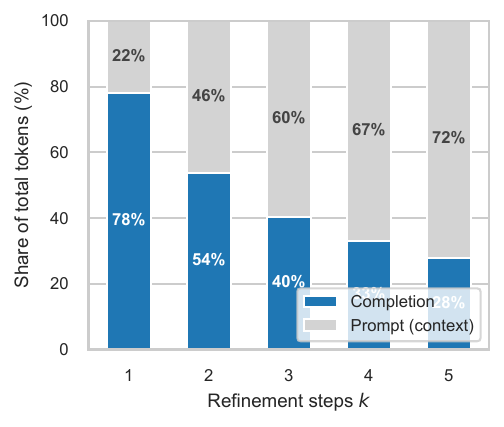}
    \caption{Study~1: share of prompt vs.\ completion tokens per refinement
             step $k$ (GPT-4.1-mini, average of L3--4 and L4--5). At $k=1$
             prompts are 22\% of all tokens; by $k=5$ they are 73\%.}
  \end{subfigure}
  \hfill
  \begin{subfigure}[t]{0.48\linewidth}
    \includegraphics[width=\linewidth]{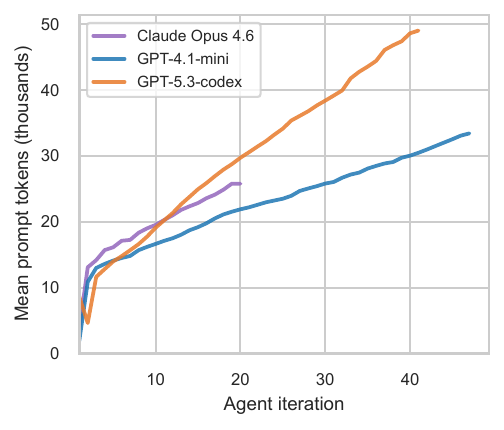}
    \caption{Study~2: mean prompt token count per agent iteration for each
             model. Lines end where fewer than half of instances remain
             active. Prompt tokens exceed 99\% of all tokens at every
             iteration.}
  \end{subfigure}
  \caption{Token composition in both studies. Since our current carbon scenario charges only for
           completion tokens, the carbon estimate covers a shrinking fraction of
           actual token workload as refinement depth or agent iterations grow.}
  \label{fig:tokens}
\end{figure}

\end{document}